\documentclass[runningheads]{llncs}
\usepackage[T1]{fontenc}
\usepackage{graphicx}
\usepackage{amsmath}
\usepackage{marginnote}
\usepackage{bbding}
\usepackage{cite}
\usepackage[rgb,x11names]{xcolor}
\usepackage{amssymb}
\usepackage{amsmath}
\usepackage{subfig}
\usepackage{hyperref}
\hypersetup{colorlinks=true,citecolor=blue}
\usepackage{graphicx}
\usepackage[]{algorithm2e}
\usepackage{mathrsfs}
\usepackage{bbding}
\usepackage{rotating}
\usepackage{multirow}
\usepackage{marginnote}
\usepackage{overpic}
\usepackage{colortbl}
\usepackage[labelfont=bf]{caption}
\usepackage{bbding}
\usepackage{tabularx}
\usepackage[marginal]{footmisc}

\begin{document}
\title{ODC-SA Net: Orthogonal Direction Enhancement and Scale Aware Network for Polyp Segmentation}
%
%

\author{Chenhao Xu\inst{*}\and
Yudian Zhang\inst{*}\and
Kaiye Xu\inst{}\and
Haijiang Zhu\inst{(}\Envelope\inst{)}}
\authorrunning{Xu et al.}
\titlerunning{Orthogonal Direction Enhancement and Scale Aware Network}%
%
\institute{CIST, Beijing University of Chemical Technology, Beijing 100029, China \\
\email{\{xuch, zyd, xuky\}@buct.edu.cn}\\
\email{zhuhj@mail.buct.edu.cn}
}
\maketitle              
\footnote{* means the authors contributed equally to this work and should be considered co-first authors.\\}
\begin{abstract}
Accurate polyp segmentation is crucial for the early detection and prevention of colorectal cancer. However, the existing polyp detection methods sometimes ignore multi-directional features and drastic
changes in scale. To address these challenges, we design an Orthogonal Direction Enhancement and Scale Aware Network (ODC-SA Net) for polyp segmentation. The Orthogonal Direction Convolutional (ODC) block can extract multi-directional features using transposed rectangular convolution kernels through forming an orthogonal feature vector basis, which solves the issue of random feature direction changes and reduces computational load. Additionally, the Multi-scale Fusion Attention (MSFA) mechanism is proposed to emphasize scale changes in both
spatial and channel dimensions, enhancing the segmentation accuracy for polyps of varying sizes. Extraction with Re-attention Module (ERA) is used to re-combinane effective features, and Structures of Shallow Reverse Attention Mechanism (SRA) is used to enhance polyp edge with low level information. A large number of experiments conducted on public datasets have demonstrated that the performance of this model is superior to state-of-the-art methods.
\end{abstract}
\section{Introduction}
According to the data of the World Health Organization, colorectal cancer is the third most common cancer in the world. Early detection of adenomatous polyps plays a crucial role in the prevention of colorectal cancer. However, polyp detection methods that rely on colonoscopy require the physician's visual judgment and manual labeling the polyp area, which is not only time-consuming and labor-intensive, but also susceptible to subjective factors. With the rapid development of deep learning technology, neural networks are increasingly widely used in the field of medical image processing.

Among many methods, the earliest polyp image segmentation methods are relying on hand-created features including color, shape, appearance, texture, or their combination\cite{seg2010, seg2012}. Most of these methods distinguish polyps from background by training a classifier. However, in clinical practice, the characteristics of polyps will change under different conditions, and the manually created features are very limited, so it is difficult to achieve good results. With the rise of deep neural networks, progress has been made in polyp segmentation, but they can only use borders to locate the general position of polyps and cannot get the specific shape and boundary\cite{seg2017}. In the past decade, Fully Convolutional Networks (FCNs) have been widely used for semantic segmentation tasks. Akbair et al. used the improved FCN to enhance the precision by smart patch selection method and adaptive thresholding method\cite{fcn}. Since U-Net\cite{unet} has shown very excellent results in semantic segmentation tasks, some variants of U-Net, such as U-Net++\cite{unet++} and ResUNet++\cite{resunet++}, have been proposed and applied to polyp segmentation tasks and achieved relatively good results. However, the above methods only focus on the whole region of the polyp and ignore the boundary information which is extremely important for this task. To solve the above problems, Fang et al. proposed SFA-Net, a Selective Feature Aggregation network with area and boundary constraints\cite{sfa}, but it is time-consuming and prone to overfitting. Fan et al. proposed PraNet, which innovatively uses reverse attention (RA) to associate the region with the boundary information and obtains the refined polyp boundary effect\cite{pranet}. Zhou et al. enhance these hierarchical features to generate finer segmentation maps by training an independent Boundary Aggregated Module (BAM) to obtain boundary information and incorporate them into the segmentation network\cite{cfanet}. Recently, researchers pay more attention to aggregate the output information of each layer in the encoder and use low-level features to fuse high-level semantic information to correct the semantic bias caused by the excessive use of high-level features, because the low-level outputs of the encoder may contain features that the high level does not have. For example, Zhao et al. proposed MS-Net which is based on multi-scale subtraction, comparing the information difference between adjacent feature levels and equipping pyramid-shaped subtraction units with different receptive fields to obtain rich multi-scale difference information to guide decoder\cite{msnet}. Polyp-PVT takes semantic and positional information from high-level features, details from low-level features and finally fuses them together\cite{polyppvt}.

However, most of the existing baseline methods can be applied to various semantic segmentation problems. Although some of them consider the particularity of polyp segmentation problem, the following unique features are still ignored: 1) In the target image containing extremely hidden polyps, the angle of the polyp changes randomly with the change of the lens, which makes the features of the polyps no longer representative in a fixed or specific few directions. The general approach to this problem is to increase the convolution depth and change the size of the convolution kernel, which undoubtedly increases the computational burden. 2) The size of the hidden object in the polyp image varies dramatically, and most of the existing methods ignore the joint effect of scale change in space and channel. 3) Most of the existing methods ignore the importance of feature reorganization for the segmentation problem, which makes the rich semantic information obtained by the encoder cannot be fully utilized. In order to solve the above problems, we proposed the Orthogonal Direction Enhancement and Scale Aware Network (ODC-SA Net). The main contributions are as follows:

\begin{itemize}
    \item Orthogonal Direction Convolutional (ODC) block is proposed. A pair of transposed rectangular convolution kernels are used to extract the horizontal and vertical features in each channel of the deepest feature map in both directions to form an orthogonal feature vector basis. Polyp features in multiple directions are represented by a linear combination of orthogonal basis vectors for all channels, which solves the problem of random changes in target feature directions and reduces the computational burden.
    \item Multi-scale Fusion Attention (MSFA) is proposed, which includes Spatial Scalable Enhancement Mechanism (S2E), Channel Scalable Attention Mechanism (CSA) and Residual Fusion Attention (RFA). Through the dual channel of convolutional and pooling, the targets with different scales are emphasized in two dimensions of space and channel, and the shallower features are used to guide the deepest features, so as to realize the fusion of multi-scale deep semantic and position information.
    \item Extraction with Re-attention Module (ERA) is proposed. Two sets of channel and spatial attention mechanism are used to mine the deep semantic information again, and the feature reorganization and weight allocation are carried out to realize the efficient utilization of deep semantic information, prevent important details from being ignored, and fully retain the original scale features.
    \item Shallow Reverse Attention Mechanism (SRA) is proposed. The shallow output of the encoder with rich boundary information is used to guide the Reverse Attention Mechanism to strengthen the segmentation target boundary and improve the accuracy.
\end{itemize}

\begin{figure}[t]
\includegraphics[width=\textwidth]{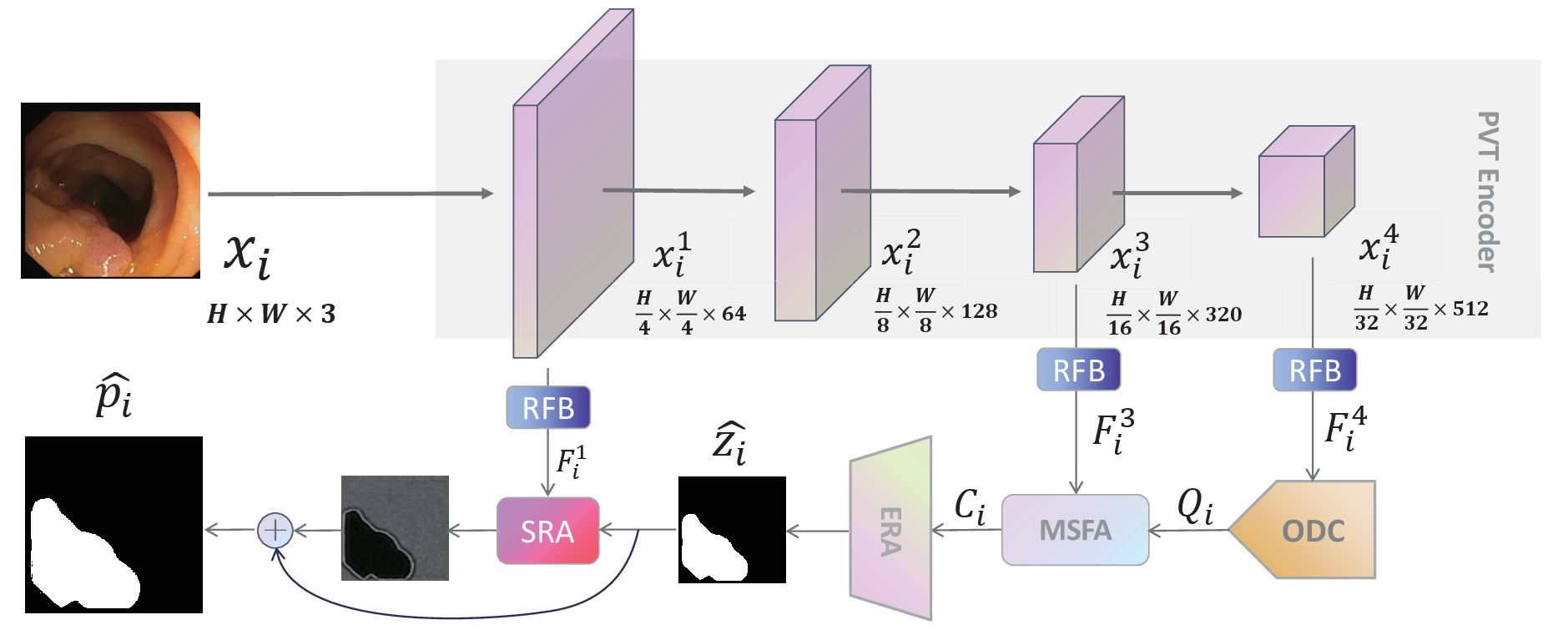}
\caption{Structure of Orthogonal Direction Enhancement and Scale Aware Network.} \label{network}
\end{figure}

\section{Methodology}

Fig. \ref{network} shows our network. The feature maps $x_i^1$, $x_i^2$ and $x_i^4$ obtained through encoding by PVTv2 encoder\cite{pvt}. They are fed into Receptive Field Block (RFB)\cite{rfb} to initially process the encoder outputs using multi-branch and dilated convolutional structure to capture a wider range of information, obtaining $F^1_i\in \mathbb{R}^{\frac{H}{4}\times \frac{W}{4} \times 32}$, $F^3_i\in \mathbb{R}^{\frac{H}{16}\times \frac{W}{16} \times 32}$ and $F^4_i\in \mathbb{R}^{\frac{H}{32}\times \frac{W}{32} \times 32}$. Then, $F^4_i$ is fed into the ODC to obtain $Q_i$. $F^3_i$ and $Q_i$ are fed into MSFA. The output $C_i$ of MSFA enters ERA, thus obtaining the preliminary segmentation result $\hat{z}_i$. Finally, SRA is used to strengthen the boundary in $\hat{z}_i$ by utilizing shallow features $F^1_i$, and obtain the final segmentation result $\hat{p}_i$.

\subsection{Orthogonal Direction Convolutional (ODC) Block }\label{fourbranch}
\subsubsection{Principle.} $x^4_i\in\mathbb{R}^{\frac{H}{32}\times \frac{W}{32} \times 512}$ has the smallest resolution and the richest semantic features due to experiencing the deepest processing in the backbone network. The most basic way to make full use of these features for extracting and synthesizing feature maps is through a series of convolutional operations. For a polyp, due to its different location and shooting direction of the colonoscopy-lens, the traditional convolution kernel will miss features in specific directions due to the rotation of the image or the target because of its square shape and fixed sliding direction. In order to compensate for this defect, the general practice is to increase the depth of the network and change the size of the convolution kernel, but these methods will undoubtedly seriously increase the computational burden and have great limitations in application. To solve this problem, we propose Orthogonal Direction Convolutional (ODC) block , in which a set of convolution kernels (1x3 and 3x1) slide along mutual perpendicular directions in the two-dimensional feature maps of $C$ channels to extract $C$ sets of orthogonal features respectively, namely $C$ sets of orthogonal basis vectors ($[\boldsymbol{c}^k, \boldsymbol{r}^k] , k\in \{1, 2,... ,C\}$). Then the features in any direction along each channel can be represented by a linear combination of them (Eq.(\ref{eq1})).

\begin{equation}
\boldsymbol{f}^k=a_k\boldsymbol{r}^k+b_k\boldsymbol{c}^k=a_kr^k\boldsymbol{i}+b_kc^k\boldsymbol{j}
\label{eq1}
\end{equation}

Then the feature map of any channel in the next layer of the network should be a combination of all the channels in the previous layer with features in any direction, namely Eq.(\ref{eq2}).

\begin{equation}
\boldsymbol{d}^k=g(\boldsymbol{f}^1,\boldsymbol{f}^2,...,\boldsymbol{f}^C)=\sum_{n=1}^{C}w_n^k\boldsymbol{f}^n=\sum_{l=1}^{C}\sum_{m=1}^{C}(\mu_{l}^k\boldsymbol{r}^l+\omega_{m}^k\boldsymbol{c}^m)
\label{eq2}
\end{equation}

In this way, we use the linear combination of orthogonal vectors to extract features in multiple directions conveniently, which solves the problem of missing features in some directions and greatly reduces the calculation amount  by replacing a large number of fully convolutional connections.

\begin{figure}[t]
\includegraphics[width=\textwidth]{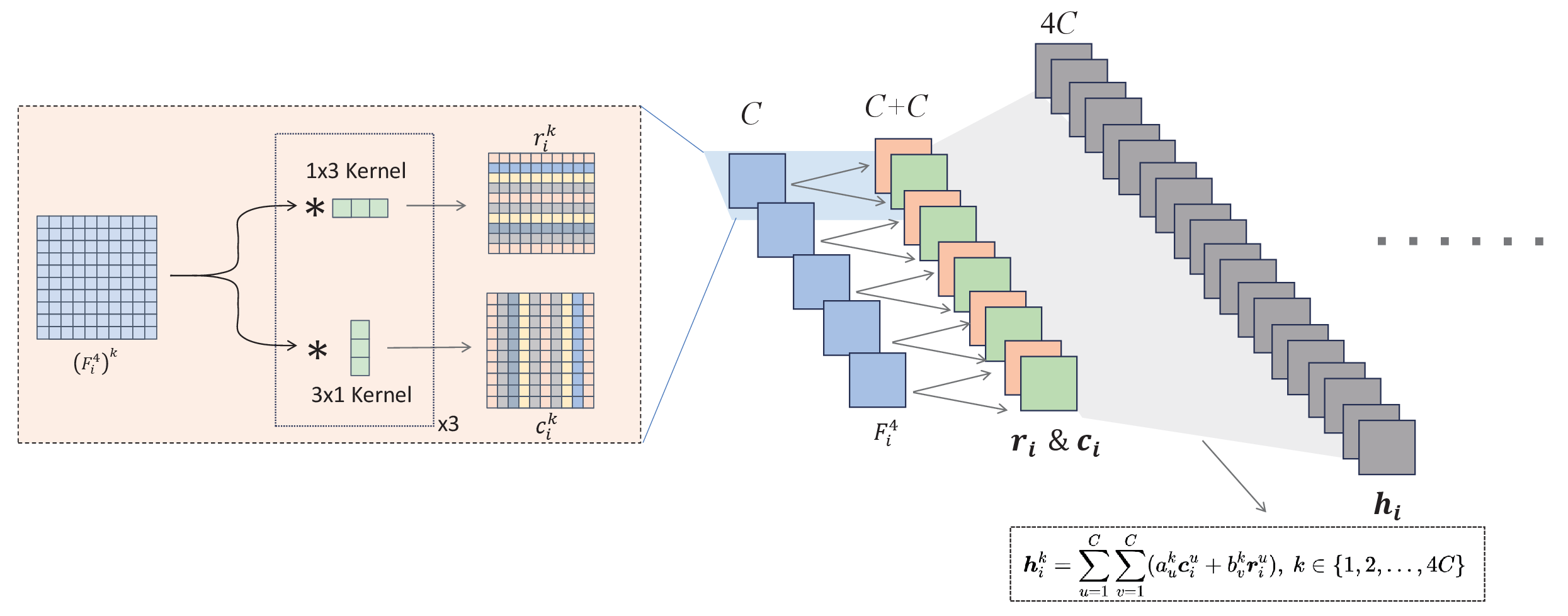}
\caption{Structure of the key sectors of Orthogonal Direction Convolutional (ODC) block, which can be also described as Eq.(\ref{eq3})-(\ref{eq5}).}\label{odc}
\end{figure}

\subsubsection{Operation.} As shown in Fig. \ref{odc}, $F^4_i\in \mathbb{R}^{\frac{H}{32}\times \frac{W}{32} \times 32}$, where the channel $C$=32, is passed into two branches: One is a 3-layer 1x3 convolution to extract horizontal features $\boldsymbol{r}_i\in \mathbb{R}^{\frac{H}{32}\times \frac{W}{32} \times 32}$, and the other is a 3-layer 3x1 convolution to extract the vertical features $\boldsymbol{c}_i\in \mathbb{R}^{\frac{H}{32}\times \frac{W}{32} \times 32}$, as shown in Eq.(\ref{eq3})(\ref{eq4}).

\begin{equation}
\boldsymbol{r}_i=(F_i^4)*\varphi_{1\times3}^\alpha*\varphi_{1\times3}^\beta*\varphi_{1\times3}^\gamma
    \label{eq3}
\end{equation} 

\begin{equation}
\boldsymbol{c}_i=(F_i^4)*\varphi_{3\times1}^\alpha*\varphi_{3\times1}^\beta*\varphi_{3\times1}^\gamma
    \label{eq4}
\end{equation} 

According to the principle of Eq.(\ref{eq2}), through the linear combination of two orthogonal feature vectors $\boldsymbol{r}_i$ and $\boldsymbol{c}_i$, the $\boldsymbol{h}_i\in \mathbb{R}^{\frac{H}{32}\times \frac{W}{32} \times 128}$ can be obtained, which contains the multi-directional features, where the number of channels is expanded by a factor of four to contain more directional features. The above operations are denoted as Eq.(\ref{eq5}).

\begin{equation}
\boldsymbol{h}_i^k=\sum_{u=1}^{C} \sum_{v=1}^{C} (a_u^k\boldsymbol{r}_i^u+b_v^k\boldsymbol{c}_i^v),\ k\in\{1,2,...,4C\}
\label{eq5}
\end{equation}

After that, we use residual connection to linearly combine the feature map $F_i^4$ with $\boldsymbol{h}_i$ in the channel dimension. The final output of this module $Q_i\in \mathbb{R}^{\frac{H}{32}\times \frac{W}{32} \times 32}$ is obtained, as in Eq.(\ref{eq6}). In fact, after the above orthogonal convolution operation, the original feature map obtains more features in all directions, and adaptively performs linear combination to achieve effective feature reorganization and fusion.

\begin{equation}
\boldsymbol{Q}_i^k=\sum_{l=1}^{4C}\sum_{m=1}^{C}[\zeta_l^k\boldsymbol{h}_i^l+\eta_m^k(F_i^4)^m]
\label{eq6}
\end{equation}

\subsection{Multi-scale Fusion Attention (MSFA)}\label{msfa}
The growth time of polyps in the intestine, the environment and the shooting distance of the colonoscopy lens are different. The polyps contained in the pictures have different sizes, and their color and texture characteristics are very similar to the background. Therefore, without increasing the number of backbone layers, if the more semantic information is fully utilized, the segmentation accuracy will be improved. Most of the existing methods treat scale-varying features as the same and ignore the attention bias caused by their differences in spatial and channel dimensions. In order to overcome the above difficulties, we propose to expand the receptive field from both spatial and channel dimensions by convolutional and pooling parallel dual channels of the high-level feature map $Q_i$ containing more detailed information, and use the feature map of the lower layer $x^3_i$, which has more global information than $Q_i$, to give attention to the deep semantics of objects with different sizes to de-emphasize the influence of background and interference.

\begin{figure}[t]
\includegraphics[width=\textwidth]{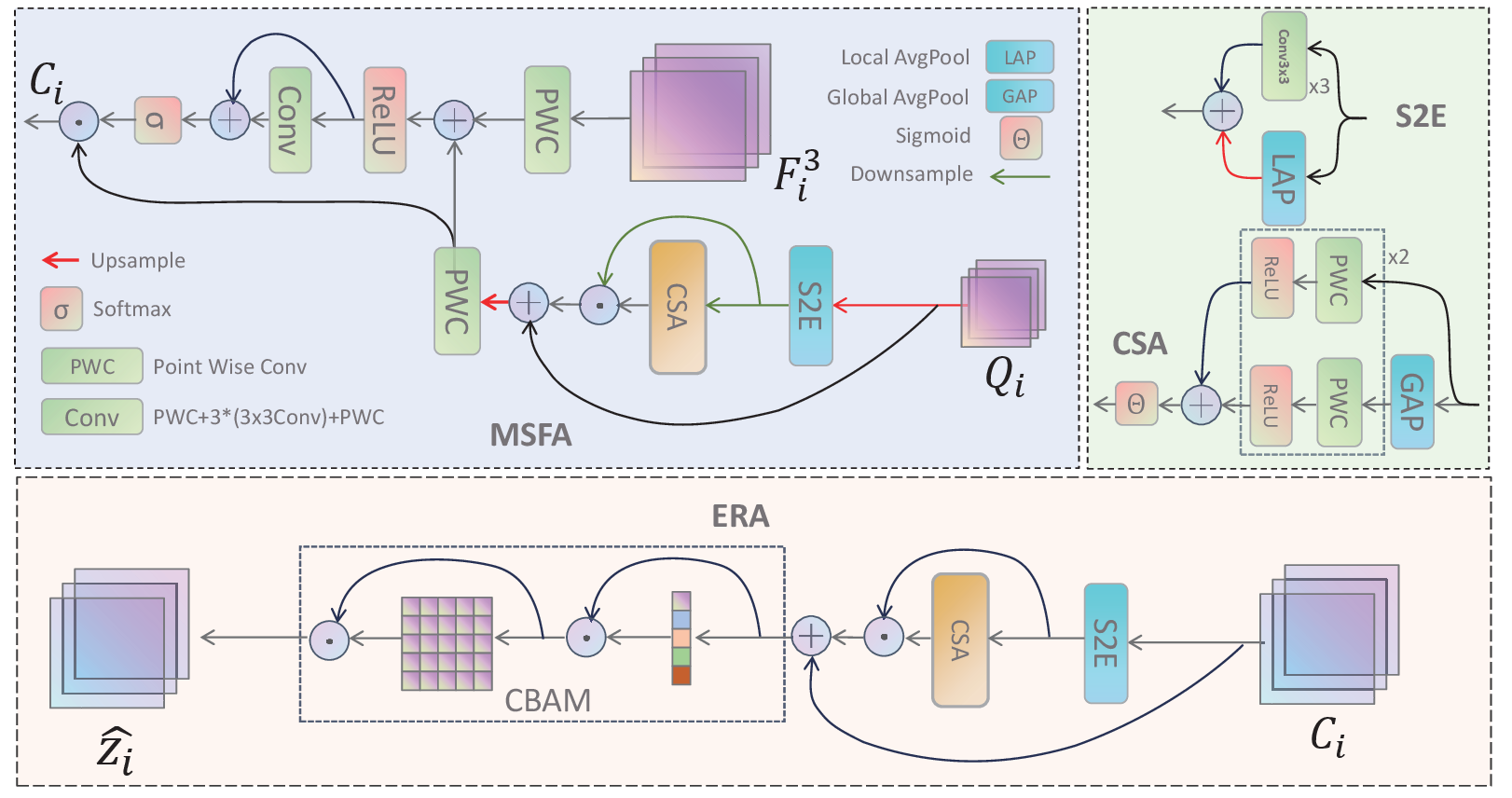}
\caption{Structures of Multi-scale Fusion Attention (MSFA) and Extraction with Re-attention Module (ERA).} \label{many}
\end{figure}

As shown in Fig. \ref{many}, firstly, $Q_i$ goes through Spatial Scalable Enhancement Mechanism (S2E) and Channel Scalable Attention Mechanism (CSA). In balancing small-scale and large-scale targets, S2E is used to reinforce multi-scale information in the spatial dimension, while CSA is used to provide multi-scale attention in the channel dimension.

\subsubsection{Spatial Scalable Enhancement Mechanism (S2E).} As shown in Fig. \ref{many}, the S2E structure contains two parallel pathways. The first pathway is composed of three 3x3 convolutional layers, which are able to capture local feature information and enhance this information through convolutional operations to focus on small objects. The second path consists of local average pooling layer: In a certain region (2x2 pixel area), since the larger target occupies most of the pixels in the region, after the average pooling operation, the features of the larger target are retained, while the small target is weakened by the average effect, that is, the overall probability distribution of the new feature map is closer to the probability distribution of the large target, so as to obtain the global context information. Therefore, the average pooling layer reduces the spatial resolution of the features by performing regional averaging operation on the input feature map, but strengthens the context global information without increasing the computational complexity of the model. After the second path is processed, the interpolation is magnified (without compensating for the smaller objects weakened in this path) to ensure the same size as the feature map obtained in the first path, and their results are added point-by-point to obtain the final output $S_i$. Such a design enables the fusion of multi-scale information in space. The above structure can adaptively obtain high-level global context information without consuming computing resources and improve the detection performance of objects with high imperceptibility.

\subsubsection{Channel Scalable Attention Mechanism (CSA).} Next, we use CSA to assign channel attention to multi-scale information. Different channels contain different semantic information, and there are scale differences in the targets mapped by these channels. In order to adaptively avoid the information loss caused by this difference, we still use the two-branch structure. As shown in Fig. \ref{many}, unlike S2E, CSA operates in the channel dimension: For small targets, the ordinary Point-Wise-Conv + ReLU structure is used to assign ordinary channel attention to them. For larger targets, global average pooling is added before the original structure, that is, the pixel average of each channel feature map is used to obtain channel attention guided by global information, which increases the weights of channels with large-scale features. Finally, the outputs of the two branches are summed and fed into Sigmoid to get the output $W_i\in(0, 1)$, which is point-wise multiplied with $Q_i$ to get $D_i$. The above operation uses S2E and CSA in series, and can make full use of multi-scale information to reduce the problem of inappropriate spatial and channel attention allocation caused by scale changes.

\subsubsection{Residual Fusion Attention (RFA).}$F^3_i$ has more global information than $D_i$, so $F^3_i$ is used to guide $D_i$ with richer semantic information to realize the fusion of deep features and shallow features. Of course, we need to upsample the $D_i$ to the same size as $F^3_i$ and adjust the channel weights by pointwise convolution before. After ReLU activation, the fusion result is entered into the \textit{Conv} block (Point-Wise-Conv + 3(3x3Conv) + Point-Wise-Conv), and the residual connection is added to accelerate the network convergence. After the above process of cross-layer feature fusion and aggregate feature extraction, the final $E_i$ generated by Softmax represents the weight of the effective information contained in each pixel, that is, the importance of the pixel in the feature map. Finally $E_i$ is multiplied point-by-point with the upsampled $D_i$ to obtain $C_i$. The above operations effectively make the model automatically focus on the target region and suppress the influence of irrelevant regions in the deep features on the segmentation results.

\subsection{Extraction with Re-attention Module (ERA)}
The $C_i$ output by MSFA has fused features from multiple layers with multiple scales. It will enter ERA to further adjust and integrate its own feature information. This process generates a feature enhancement and combination graph that is more in line with the target. It allows different features to be combined into better features and can also realize the allocation of different feature emphasis degrees. As shown in Fig. \ref{many}, ERA consists of S2E and CSA in series, which can effectively extract the features of the two dimensions of space and channel that has been discussed in section \ref{msfa}. After CSA, we added a CBAM\cite{cbam} with the same emphasis on spatial and channel attention as S2E and CSA. Using a combination of two sets of spatial and channel attention also enables a wider fitting range than when using a single module.

\begin{figure}[t]
\includegraphics[width=\textwidth]{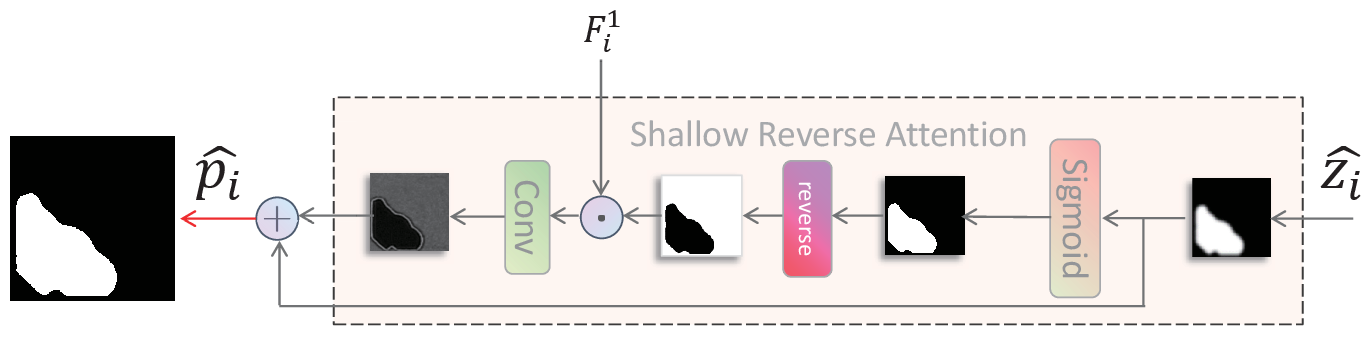}
\caption{Structures of Shallow Reverse Attention Mechanism (SRA).} \label{sra}
\end{figure}

\subsection{Shallow Reverse Attention Mechanism (SRA)}
In the polyp segmentation task, the boundary between the background and the target is not obvious, so providing the corresponding boundary information will improve the segmentation accuracy. Inspired by PraNet\cite{pranet}, we introduce Reverse Attention Module, which is confirmed to work well in mining boundary information. As shown in Fig. \ref{sra}, RA exploits the details of complementary regions by erasing the polyp regions that have been judged. But different from PraNet, which uses higher level feature maps for Reverse Attention, our network only uses the lowest level feature map to operate with the output to enhance the boundary information, which is called Shallow Reverse Attention Mechanism (SRA). This is because according to some studies\cite{edge1, edge2}, low-level feature maps contain more boundary information than high-level feature maps. In this way, the output segmentation result $\hat{z}_i$ is fused with $F_i^1$ by SRA to obtain the enhanced boundary information $F_{edge}$, which is added point-by-point with $\hat{z}_i$ to obtain a more accurate result $\hat{p}_i$.

\subsection{Loss Function}
We use the loss function $\mathcal{L}=\mathcal{L}_{BCE}^w+\mathcal{L}_{IoU}^w$ proposed by \cite{wloss}, which combines the advantages of the BCE Loss and IoU Loss, and assigns different weights according to the difference between the pixel and its surroundings, thereby guiding the network to pay attention to more detailed information. We adopt the strategy of deep supervision. The intermediate result $\hat{z}_i$ shown in Fig. \ref{network} is upsampled to $\hat{z}_i^{up}$ of the same size as the truth value $G$, and the final result $\hat{p}_i$ is supervised in this way. The loss function can be expressed as Eq.(\ref{eq7}).

\begin{equation}
\mathcal{L}_{total}=\mathcal{L}(G,\hat{p}_i)+ \mathcal{L}(G,\hat{z}_i^{up})
\label{eq7}
\end{equation}

\begin{figure}[t]
\includegraphics[width=\textwidth]{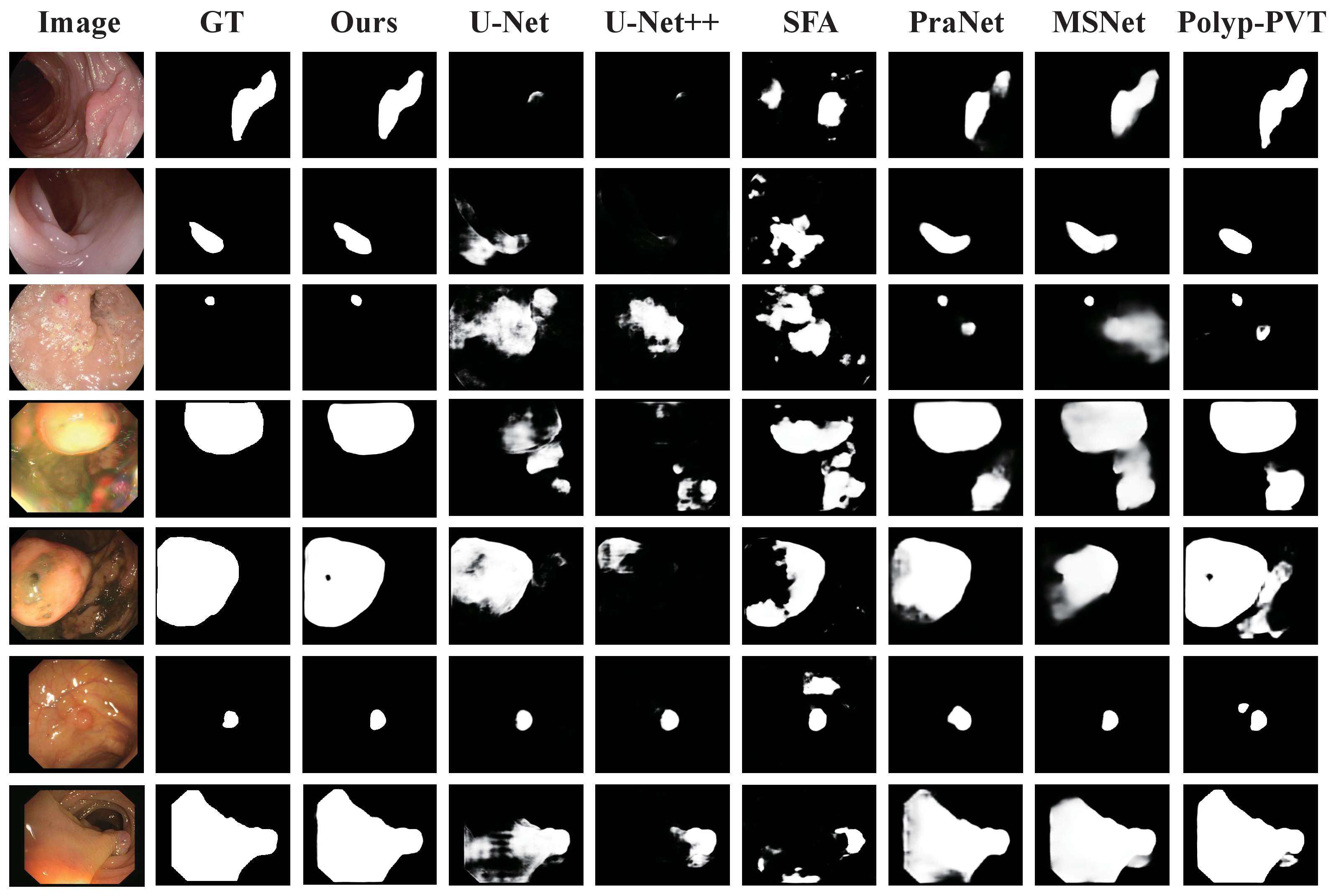}
\caption{Qualitative results of different methods.} \label{result}.
\end{figure}

\section{Experiment}
\subsection{Details}
\subsubsection{Dataset.} The dataset is same as the article \cite{pranet}, which has been widely used for training and model evaluation of polyp segmentation tasks. The dataset consists of five publicly available datasets: Kvasir\cite{kvasir}, CVC-ClinicDB\cite{cvccli}, CVC-300\cite{cvc300}, CVC-ColonDB\cite{cvccol} and ETIS-LaribPolypDB\cite{etis}.

\subsubsection{Training environment and settings.} We use PyTorch on a single 16GB NVIDIA-V100 GPU. All input images are resized to 352x352. We carry out a multi-scale training strategy \{0.75, 1, 1.25\} instead of data augmentation. We employ the AdaX\cite{adax} optimizer with an initial learning rate of 1e-4 and a 10-fold reduction in the learning rate every 30 epochs of training. A total of 100 epochs were trained with a batchsize of 16 each.

\subsubsection{Evaluation metrics.} We adopted the commonly used mean Dice (mDice), mean IoU (mIoU) and other four metrics: $S_\alpha$\cite{sa} to evaluate the similarity of the results to the true value; $F_\beta^w$\cite{fwb} to fix Dice's equal-importance flaws; $E_\phi^{max}$\cite{ef} to reflect the matching degree of results in global and detail; MAE\cite{mae} to evaluate the accuracy of the prediction results at the pixel level.

\begin{table}[!ht]
\caption{Quantitative metrics results for 7 different methods on each dataset.}\label{tab1}
\centering
\begin{tabular}{|c|l|m{1.2cm}m{1.2cm}m{1.2cm}m{1.2cm}m{1.2cm}m{1.2cm}|} 
\hline
Datasets & Methods                & mDice          & mIoU           & $F_\beta^w$        & $S_\alpha$ & $E_\phi^{max}$ & MAE             \\ 
\hline
         & U-Net(MICCAI'15)       & 0.823          & 0.755          & 0.811          & 0.889                        & 0.954                                                    & 0.019           \\
         & U-Net++(DLMIA'18)      & 0.794          & 0.729          & 0.785          & 0.873                        & 0.931                                                    & 0.022           \\
         & SFA(MICCAI'19)         & 0.700          & 0.607          & 0.647          & 0.793                        & 0.885                                                    & 0.042           \\
ClinicDB & PraNet(MICCAI'20)      & 0.899          & 0.849          & 0.896          & 0.936                        & 0.979                                                    & 0.009           \\
         & MSNet(MICCAI'21)       & 0.918          & 0.869          & 0.913          & 0.946                        & 0.979                                                    & 0.008           \\
         & Polyp-PVT(CAAI AIR'23) & 0.937          & 0.889          & \textbf{0.936} & 0.949                        & 0.985                                                    & 0.006           \\
         & OURS                   & \textbf{0.943} & \textbf{0.899} & 0.934          & \textbf{0.952}               & \textbf{0.986}                                           & \textbf{0.006}  \\ 
\hline
         & U-Net(MICCAI'15)       & 0.818          & 0.746          & 0.794          & 0.858                        & 0.893                                                    & 0.055           \\
         & U-Net++(DLMIA'18)      & 0.821          & 0.743          & 0.808          & 0.862                        & 0.910                                                    & 0.048           \\
         & SFA(MICCAI'19)         & 0.723          & 0.611          & 0.670          & 0.782                        & 0.849                                                    & 0.075           \\
Kvasir   & PraNet(MICCAI'20)      & 0.898          & 0.840          & 0.885          & 0.915                        & 0.948                                                    & 0.030           \\
         & MSNet(MICCAI'21)       & 0.905          & 0.849          & 0.892          & 0.923                        & 0.954                                                    & 0.028           \\
         & Polyp-PVT(CAAI AIR'23) & 0.917          & 0.864          & 0.911          & 0.925                        & 0.956                                                    & \textbf{0.023}  \\
         & OURS                   & \textbf{0.921} & \textbf{0.875} & \textbf{0.915} & \textbf{0.927}               & \textbf{0.967}                                           & \textbf{0.023}  \\ 
\hline
         & U-Net(MICCAI'15)       & 0.398          & 0.335          & 0.366          & 0.684                        & 0.740                                                    & 0.036           \\
         & U-Net++(DLMIA'18)      & 0.401          & 0.344          & 0.390          & 0.683                        & 0.766                                                    & 0.035           \\
         & SFA(MICCAI'19)         & 0.297          & 0.217          & 0.231          & 0.557                        & 0.633                                                    & 0.109           \\
ETIS     & PraNet(MICCAI'20)      & 0.628          & 0.567          & 0.600          & 0.794                        & 0.841                                                    & 0.031           \\
         & MSNet(MICCAI'21)       & 0.723          & 0.652          & 0.677          & 0.845                        & 0.890                                                    & 0.020           \\
         & Polyp-PVT(CAAI AIR'23) & 0.787          & 0.706          & 0.750          & 0.871                        & \textbf{0.906}                                           & \textbf{0.013}  \\
         & OURS                   & \textbf{0.803} & \textbf{0.728} & \textbf{0.765} & \textbf{0.876}               & \textbf{0.906}                                           & \textbf{0.013}  \\ 
\hline
         & U-Net(MICCAI'15)       & 0.710          & 0.627          & 0.684          & 0.843                        & 0.876                                                    & 0.022           \\
         & U-Net++(DLMIA'18)      & 0.707          & 0.624          & 0.687          & 0.839                        & 0.898                                                    & 0.018           \\
         & SFA(MICCAI'19)         & 0.467          & 0.329          & 0.341          & 0.640                        & 0.817                                                    & 0.065           \\
CVC-300  & PraNet(MICCAI'20)      & 0.871          & 0.797          & 0.843          & 0.925                        & 0.972                                                    & 0.010           \\
         & MSNet(MICCAI'21)       & 0.865          & 0.799          & 0.848          & 0.926                        & 0.953                                                    & 0.010           \\
         & Polyp-PVT(CAAI AIR'23) & 0.900          & 0.833          & \textbf{0.884} & \textbf{0.935}               & \textbf{0.973}                                           & \textbf{0.007}  \\
         & OURS                   & \textbf{0.902} & \textbf{0.836} & 0.880          & \textbf{0.935}               & 0.967                                                    & \textbf{0.007}  \\ 
\hline
         & U-Net(MICCAI'15)       & 0.512          & 0.444          & 0.498          & 0.712                        & 0.776                                                    & 0.061           \\
         & U-Net++(DLMIA'18)      & 0.483          & 0.410          & 0.467          & 0.691                        & 0.760                                                    & 0.064           \\
         & SFA(MICCAI'19)         & 0.469          & 0.347          & 0.379          & 0.634                        & 0.765                                                    & 0.094           \\
ColonDB  & PraNet(MICCAI'20)      & 0.709          & 0.640          & 0.696          & 0.819                        & 0.869                                                    & 0.045           \\
         & MSNet(MICCAI'21)       & 0.751          & 0.671          & 0.736          & 0.838                        & 0.883                                                    & 0.041           \\
         & Polyp-PVT(CAAI AIR'23) & 0.808          & 0.727          & \textbf{0.795} & \textbf{0.865}               & 0.913                                                    & 0.031           \\
         & OURS                   & \textbf{0.810} & \textbf{0.730} & 0.792          & 0.863                        & \textbf{0.916}                                           & \textbf{0.028}  \\
\hline
\end{tabular}
\end{table}

\subsection{Comparison Experiment}\label{ce}
We compare our model with SOTAs both quantitatively and qualitatively, including: U-Net\cite{unet}, U-Net++\cite{unet++}, SFA\cite{sfa}, PraNet\cite{pranet}, MSNet\cite{msnet} and Polyp-PVT\cite{polyppvt}. One part of the CVC-ClinicDB and Kvasir datasets are used for training and the other for testing. CVC-300, ETIS-LaribPolypDB, and CVC-ColonDB are all used as test datasets. Table \ref{tab1} shows that the model outperforms all SOTAs on these Datasets. Fig. \ref{result} shows some prediction results of different models.

\begin{figure}[t]
\includegraphics[width=\textwidth]{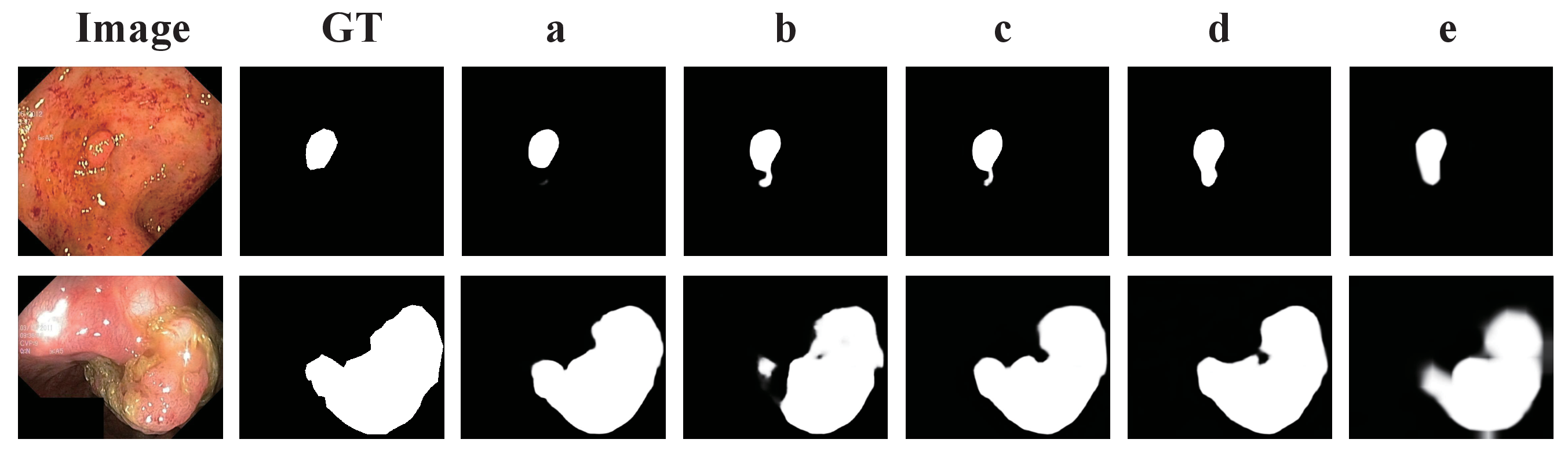}
\caption{Qualitative results of ablation experiments.} \label{abr}.
\end{figure}

\subsection{Ablation Experiments}
To better understand the role of each component of our model, we performed ablation experiments. Table \ref{tab2} lists the experiment results of the two datasets with the highest and lowest average accuracy. CVC-ClinicDB is a representative of the test dataset that is similar to the training dataset, while ETIS-LaribPolypDB is a typical unseen test dataset. Fig. \ref{abr} shows some prediction results with different components. When reducing any of the proposed components, the predicted mDice, mIoU and $F_\beta^w$ decrease, which is enough to illustrate the effectiveness of the components.

\begin{figure}[t]
\includegraphics[width=\textwidth]{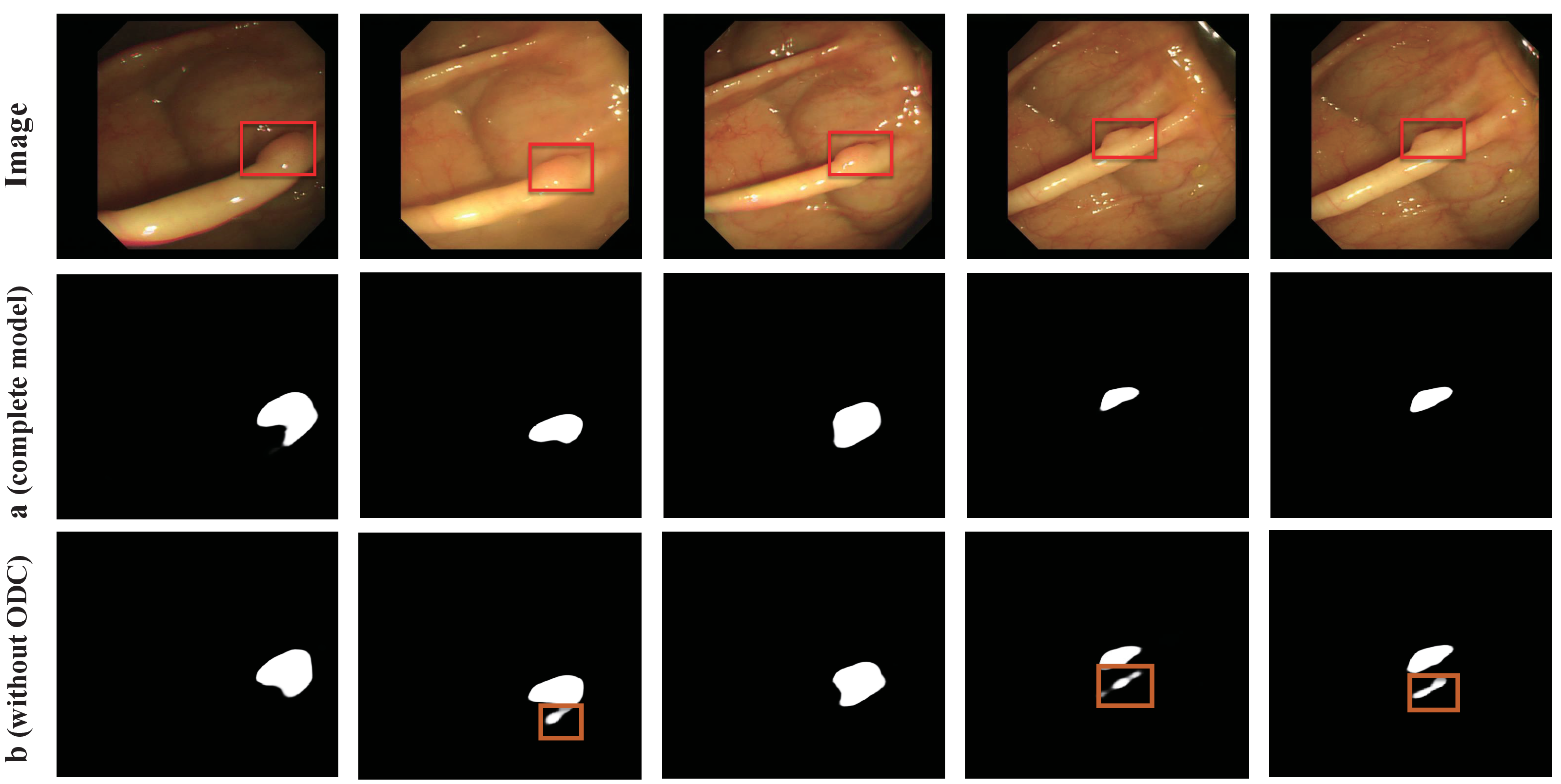}
\caption{Results for an example of polyp direction moving with the camera.} \label{abr2}.
\end{figure}

In particular, the important role of ODC is analyzed in the following. Quantitatively, as shown in Table \ref{tab2}, when ODC was added to the model, the mDice on ETIS-LaribPolypDB significantly increased from 0.788 to 0.803. Qualitatively, general datasets contain multiple sets of consecutive colonoscopy images, which contain the same polyp target at different angles while the camera moves. Network containing ODC shows the advantage when the target features are in different directions. For example, Fig. \ref{abr2} shows a set of sequentially taken images. It can be clearly seen that when the direction of the polyp target is deviated, the model without ODC may have cognitive bias towards the same target at different angles, while the complete model with ODC overcomes this problem.

\begin{table}
\caption{Quantitative metrics results for ablation experiments with different components.}\label{tab2}
\centering
\begin{tabular}{|c|ccccc|m{1.2cm}m{1.2cm}m{1.2cm}|m{1.2cm}m{1.2cm}m{1.2cm}|} 
\hline
  &           &           &           &           &           &                & ClinicDB       &                         &                & ETIS           &        \\ 
\cline{7-12}
  & Backbone  & ODC       & SRA       & ERA       & MSFA      & mDice          & mIoU           & $F_\beta^w$                     & mDice          & mIoU           & $F_\beta^w$    \\ 
\hline
a & \checkmark & \checkmark & \checkmark & \checkmark & \checkmark & \textbf{0.943} & \textbf{0.899} & \textbf{0.934}          & \textbf{0.803} & \textbf{0.728} & \textbf{0.765}  \\
b &\checkmark &           & \checkmark & \checkmark & \checkmark & 0.937          & 0.893          & \textbf{\textbf{0.934}} & 0.788          & 0.712          & 0.749  \\
c & \checkmark &           &           & \checkmark & \checkmark & 0.922          & 0.877          & 0.915                   & 0.780          & 0.700          & 0.733  \\
d & \checkmark &           &           &           & \checkmark & 0.914          & 0.875          & 0.900                   & 0.776          & 0.706          & 0.735  \\
e & \checkmark &           &           &           &           & 0.903          & 0.861          & 0.889                   & 0.748          & 0.671          & 0.691  \\
\hline
\end{tabular}
\end{table}

\section{Conclusion}
In this work, we propose the Orthogonal Direction Enhancement and Scale Aware Network for Polyp Segmentation, which consists of Orthogonal Direction Convolutional block, Multi-scale Fusion Attention, Extraction with Re-attention Module and Shallow Reverse Attention Mechanism. The overall segmentation results are better than other methods quantitatively and qualitatively.

\subsubsection{Acknowledgements.} Research supported by the Training Program of Innovation and Entrepreneurship for Undergraduates in Beijing University of Chemical Technology (X202410010085).

%
%
%
\bibliographystyle{splncs04}
\bibliography{ref}

\begin{thebibliography}{10}
\providecommand{\url}[1]{\texttt{#1}}
\providecommand{\urlprefix}{URL }
\providecommand{\doi}[1]{https://doi.org/#1}

\bibitem{fwb}
Achanta, R., Hemami, S., Estrada, F., Susstrunk, S.: Frequency-tuned salient region detection. In: 2009 IEEE conference on computer vision and pattern recognition. pp. 1597--1604. IEEE (2009)

\bibitem{fcn}
Akbari, M., Mohrekesh, M., Nasr-Esfahani, E., Soroushmehr, S.R., Karimi, N., Samavi, S., Najarian, K.: Polyp segmentation in colonoscopy images using fully convolutional network. In: 2018 40th Annual International Conference of the IEEE Engineering in Medicine and Biology Society (EMBC). pp. 69--72. IEEE (2018)

\bibitem{cvccli}
Bernal, J., S{\'a}nchez, F.J., Fern{\'a}ndez-Esparrach, G., Gil, D., Rodr{\'\i}guez, C., Vilari{\~n}o, F.: Wm-dova maps for accurate polyp highlighting in colonoscopy: Validation vs. saliency maps from physicians. Computerized medical imaging and graphics  \textbf{43},  99--111 (2015)

\bibitem{polyppvt}
Bo, D., Wenhai, W., Deng-Ping, F., Jinpeng, L., Huazhu, F., Ling, S.: Polyp-pvt: Polyp segmentation with pyramidvision transformers (2023)

\bibitem{sa}
Fan, D.P., Cheng, M.M., Liu, Y., Li, T., Borji, A.: Structure-measure: A new way to evaluate foreground maps. In: Proceedings of the IEEE international conference on computer vision. pp. 4548--4557 (2017)

\bibitem{ef}
Fan, D.P., Gong, C., Cao, Y., Ren, B., Cheng, M.M., Borji, A.: Enhanced-alignment measure for binary foreground map evaluation (2018)

\bibitem{pranet}
Fan, D.P., Ji, G.P., Zhou, T., Chen, G., Fu, H., Shen, J., Shao, L.: Pranet: Parallel reverse attention network for polyp segmentation. In: International conference on medical image computing and computer-assisted intervention. pp. 263--273. Springer (2020)

\bibitem{edge1}
Fan, D.P., Zhou, T., Ji, G.P., Zhou, Y., Chen, G., Fu, H., Shen, J., Shao, L.: Inf-net: Automatic covid-19 lung infection segmentation from ct images. IEEE transactions on medical imaging  \textbf{39}(8),  2626--2637 (2020)

\bibitem{sfa}
Fang, Y., Chen, C., Yuan, Y., Tong, K.y.: Selective feature aggregation network with area-boundary constraints for polyp segmentation. In: Medical Image Computing and Computer Assisted Intervention--MICCAI 2019: 22nd International Conference, Shenzhen, China, October 13--17, 2019, Proceedings, Part I 22. pp. 302--310. Springer (2019)

\bibitem{resunet++}
Jha, D., Smedsrud, P.H., Riegler, M.A., Johansen, D., De~Lange, T., Halvorsen, P., Johansen, H.D.: Resunet++: An advanced architecture for medical image segmentation. In: 2019 IEEE international symposium on multimedia (ISM). pp. 225--2255. IEEE (2019)

\bibitem{adax}
Li, W., Zhang, Z., Wang, X., Luo, P.: Adax: Adaptive gradient descent with exponential long term memory. arXiv preprint arXiv:2004.09740  (2020)

\bibitem{rfb}
Liu, S., Huang, D., et~al.: Receptive field block net for accurate and fast object detection. In: Proceedings of the European conference on computer vision (ECCV). pp. 385--400 (2018)

\bibitem{seg2012}
Park, S.Y., Sargent, D., Spofford, I., Vosburgh, K.G., Yousif, A., et~al.: A colon video analysis framework for polyp detection. IEEE Transactions on Biomedical Engineering  \textbf{59}(5),  1408--1418 (2012)

\bibitem{mae}
Perazzi, F., Kr{\"a}henb{\"u}hl, P., Pritch, Y., Hornung, A.: Saliency filters: Contrast based filtering for salient region detection. In: 2012 IEEE conference on computer vision and pattern recognition. pp. 733--740. IEEE (2012)

\bibitem{kvasir}
Pogorelov, K., Randel, K.R., Griwodz, C., Eskeland, S.L., de~Lange, T., Johansen, D., Spampinato, C., Dang-Nguyen, D.T., Lux, M., Schmidt, P.T., et~al.: Kvasir: A multi-class image dataset for computer aided gastrointestinal disease detection. In: Proceedings of the 8th ACM on Multimedia Systems Conference. pp. 164--169 (2017)

\bibitem{unet}
Ronneberger, O., Fischer, P., Brox, T.: U-net: Convolutional networks for biomedical image segmentation. In: Medical image computing and computer-assisted intervention--MICCAI 2015: 18th international conference, Munich, Germany, October 5-9, 2015, proceedings, part III 18. pp. 234--241. Springer (2015)

\bibitem{etis}
Silva, J., Histace, A., Romain, O., Dray, X., Granado, B.: Toward embedded detection of polyps in wce images for early diagnosis of colorectal cancer. International journal of computer assisted radiology and surgery  \textbf{9},  283--293 (2014)

\bibitem{cvccol}
Tajbakhsh, N., Gurudu, S.R., Liang, J.: Automated polyp detection in colonoscopy videos using shape and context information. IEEE transactions on medical imaging  \textbf{35}(2),  630--644 (2015)

\bibitem{seg2010}
Van~Wijk, C., Van~Ravesteijn, V.F., Vos, F.M., Van~Vliet, L.J.: Detection and segmentation of colonic polyps on implicit isosurfaces by second principal curvature flow. IEEE Transactions on Medical Imaging  \textbf{29}(3),  688--698 (2010)

\bibitem{cvc300}
V{\'a}zquez, D., Bernal, J., S{\'a}nchez, F.J., Fern{\'a}ndez-Esparrach, G., L{\'o}pez, A.M., Romero, A., Drozdzal, M., Courville, A.: A benchmark for endoluminal scene segmentation of colonoscopy images. Journal of healthcare engineering  \textbf{2017} (2017)

\bibitem{pvt}
Wang, W., Xie, E., Li, X., Fan, D.P., Song, K., Liang, D., Lu, T., Luo, P., Shao, L.: Pyramid vision transformer: A versatile backbone for dense prediction without convolutions. In: Proceedings of the IEEE/CVF international conference on computer vision. pp. 568--578 (2021)

\bibitem{wloss}
Wei, J., Wang, S., Huang, Q.: F$^3$net: fusion, feedback and focus for salient object detection. In: Proceedings of the AAAI conference on artificial intelligence. vol.~34, pp. 12321--12328 (2020)

\bibitem{cbam}
Woo, S., Park, J., Lee, J.Y., Kweon, I.S.: Cbam: Convolutional block attention module. In: Proceedings of the European conference on computer vision (ECCV). pp. 3--19 (2018)

\bibitem{seg2017}
Yu, L., Chen, H., Dou, Q., Qin, J., Heng, P.A.: Integrating online and offline three-dimensional deep learning for automated polyp detection in colonoscopy videos. IEEE journal of biomedical and health informatics  \textbf{21}(1),  65--75 (2016)

\bibitem{edge2}
Zhang, Z., Fu, H., Dai, H., Shen, J., Pang, Y., Shao, L.: Et-net: A generic edge-attention guidance network for medical image segmentation. In: Medical Image Computing and Computer Assisted Intervention--MICCAI 2019: 22nd International Conference, Shenzhen, China, October 13--17, 2019, Proceedings, Part I 22. pp. 442--450. Springer (2019)

\bibitem{msnet}
Zhao, X., Zhang, L., Lu, H.: Automatic polyp segmentation via multi-scale subtraction network. In: Medical Image Computing and Computer Assisted Intervention--MICCAI 2021: 24th International Conference, Strasbourg, France, September 27--October 1, 2021, Proceedings, Part I 24. pp. 120--130. Springer (2021)

\bibitem{cfanet}
Zhou, T., Zhou, Y., He, K., Gong, C., Yang, J., Fu, H., Shen, D.: Cross-level feature aggregation network for polyp segmentation. Pattern Recognition  \textbf{140},  109555 (2023)

\bibitem{unet++}
Zhou, Z., Rahman~Siddiquee, M.M., Tajbakhsh, N., Liang, J.: Unet++: A nested u-net architecture for medical image segmentation. In: Deep Learning in Medical Image Analysis and Multimodal Learning for Clinical Decision Support: 4th International Workshop, DLMIA 2018, and 8th International Workshop, ML-CDS 2018, Held in Conjunction with MICCAI 2018, Granada, Spain, September 20, 2018, Proceedings 4. pp. 3--11. Springer (2018)

\end{thebibliography}

\end{document}